\documentclass[sigconf]{acmart}

\usepackage{amssymb}
\usepackage{subfigure}
\usepackage{booktabs}

\usepackage{breakurl}

\AtBeginDocument{%
  \providecommand\BibTeX{{%
    \normalfont B\kern-0.5em{\scshape i\kern-0.25em b}\kern-0.8em\TeX}}}

\setcopyright{none}
\renewcommand\footnotetextcopyrightpermission[1]{}

\newcommand\blfootnote[1]{%
  \begingroup
  \renewcommand\thefootnote{}\footnote{#1}%
  \addtocounter{footnote}{-1}%
  \endgroup
}

\begin{document}

\fancyhead{}

\title{Cracking the Black Box: Distilling Deep Sports Analytics}

\author{Xiangyu Sun, Jack Davis, Oliver Schulte, Guiliang Liu}
\affiliation{%
  \institution{Simon Fraser University\\
8888 University Dr, Burnaby, BC, Canada\\
\{xiangyu\_sun, jackd, gla68\}@sfu.ca, oschulte@cs.sfu.ca
  }
}

\begin{abstract}
This paper addresses the trade-off between Accuracy and Transparency for deep learning applied to sports analytics. Neural nets achieve great predictive accuracy through deep learning, and are popular in sports analytics \cite{fernandez2019decomposing, DBLP:conf/ijcai/LiuS18, burke2019deepqb, wang2018advantage}. But it is hard to interpret a neural net model and harder still to extract actionable insights from the knowledge implicit in it. Therefore, we built a simple and transparent model that mimics the output of the original deep learning model and represents the learned knowledge in an explicit interpretable way. Our mimic model is a linear model tree, which combines a collection of linear models with a regression-tree structure. The tree version of a neural network achieves high fidelity, explains itself, and produces insights for expert stakeholders such as athletes and coaches. We propose and compare several scalable model tree learning heuristics to address the computational challenge from datasets with millions of data points.
\end{abstract}

\maketitle

\section{Introduction}
Both neural networks and tree-based method are widely used in machine learning and sports analytics \cite{DBLP:conf/ijcai/LiuS18, fernandez2019decomposing, burke2019deepqb, wang2018advantage} to obtain actionable information. They can provide predictions for not just hypothetical situations but counterfactual ones as well. If one is using either method to estimate the chance of a shot in ice hockey or soccer resulting in a goal, and that method uses variables like ``distance to the net", ``number of players between the shooter than the net", and ``type of shot", then one can use the model to ask questions like ``what happens if the shooter performs a chip shot instead of a standard shot?" or ``how much greater is the success chance if I cut the distance to the net by half?". There are two operative differences in these predictions between trees and neural networks, predictive accuracy and transparency. \blfootnote{Accepted by the 26th ACM SIGKDD Conference on Knowledge Discovery and Data Mining (KDD 2020)}

Without introducing a great deal of complexity, trees are weak classifiers and regressors; they leave a lot of variance unexplained or cases mis-classified. When there is enough complexity for a tree to make good regressions or classifications, the resultant tree over overfits the data it was trained on. Furthermore, trees can be sensitive to small changes in the training data. This instability is serious enough that trees are rarely taken alone and instead are used in random forests \cite{ho1995random}, which are ensembles of trees in which each tree is trained on a subset of the variables and observations available. By comparison, neural networks are strong predictors. They typically produce predictive values that are much closer to reality, even on new, similar, observations that weren't part of the training data. Neural networks are much better than trees in terms of output quality.
% Furthermore, neural networks scale better in their ability to predict multiple variables or complex structures of variables all at once - for trees it's usually better to make a new tree for each desired response variable. In short, networks are much better than trees in terms of output quality.

Transparency is the other major difference: to apply predictions to a tree, simply start at the top of the tree and apply the decision rules until a leaf is reached. It is clear why a particular prediction was made, and what variables contributed to that prediction. Counterfactual predictions can be applied in the same way. Therefore, a tree model is transparent. By contrast, neural networks are black boxes: there is no clear path between any one variable and its effect on predictions. After a couple of intermediate layers of neurons, every input variable can have a non-trivial and non-obvious effect on the output. To explore counter-factual possibilities, it is necessary to run a set of variable values through the entire neural network, rather than examine any small piece. Neural networks are opaque.

This tradeoff between accuracy and transparency poses a major problem in sports analytics. We are often confronted with a great deal of variables and observations from which we need to make high quality predictions, and yet we need to make these predictions in such a way that it is clear which variables need to be manipulated in order to increase a team or single athlete's success. 

\begin{figure*}[t]
  \centering
  \includegraphics[width=\linewidth]{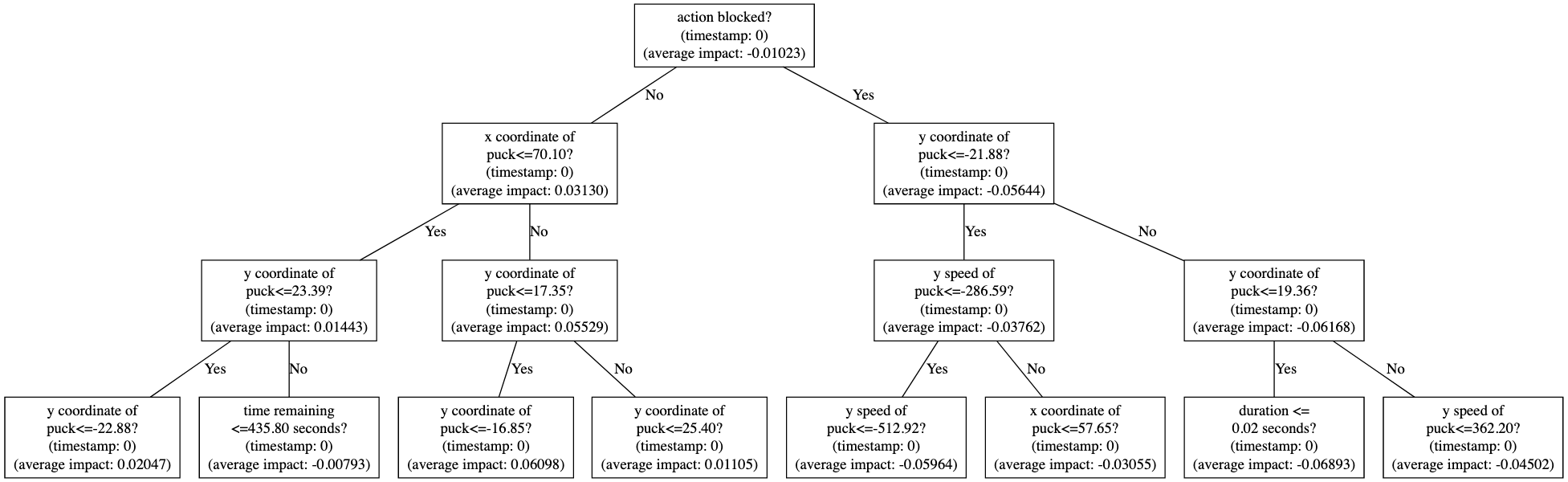}
  \caption{Model Tree Example With 4 Layers for Impact of Shots in Ice Hockey}
  \label{fig:Model Tree Example with 4 Layers Ice Hockey}
\end{figure*}

\begin{figure*}[t]
  \subfigure{
  \includegraphics[width=0.45\linewidth]{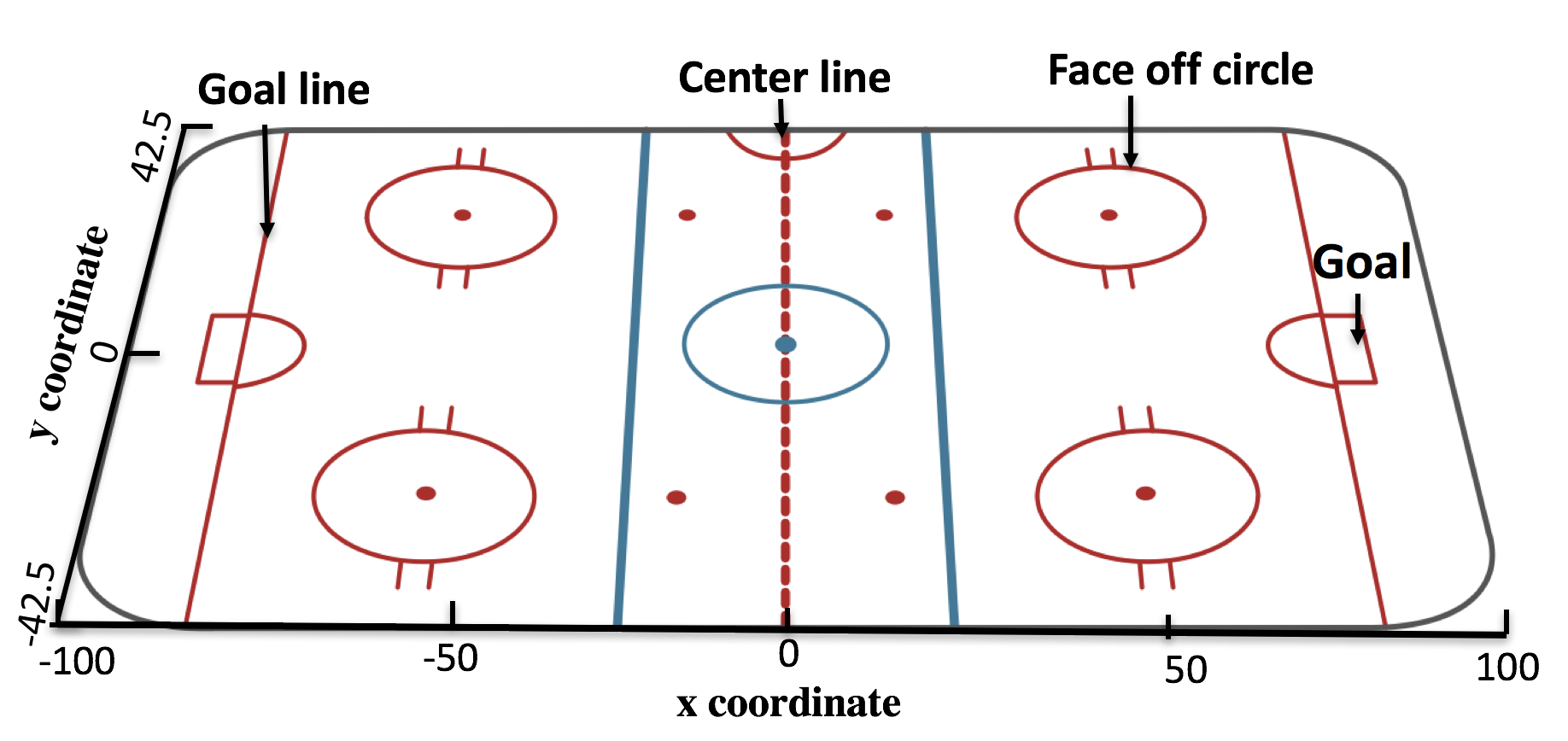}
  \hfill
  \includegraphics[width=0.45\linewidth]{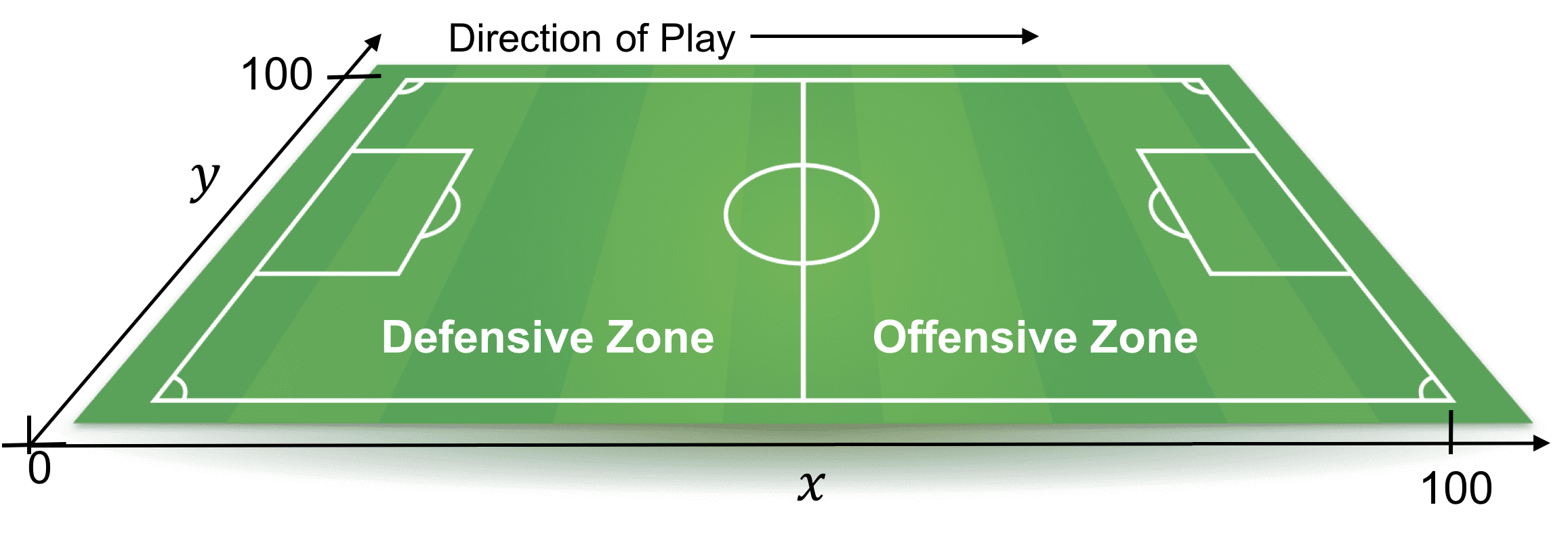}
  }
  \caption{Ice Hockey Rink and Soccer Field With Coordinates}
  \label{fig:Ice Hockey Rink and Soccer Field}
\end{figure*}

Mimic Learning is an approach that aims to get the best of both worlds: transparency without sacrificing an acceptable degree of accuracy. The basic idea is to learn an accurate black-box model, like a deep reinforcement learning (DRL) model with neural net, then train a transparent white-box model, like a regression tree, to mimic the black-box model, thereby inheriting much of the predictive accuracy. Mimic learning with tree models can be seen as knowledge extraction from a trained neural net: The tree thresholds on predictive features represent critical values for predicting response variable. It is easy to compute a feature importance metric from the tree. This informs the user which features are most influential for the neural network predictions. Finally, a mimic tree extracts rules as if-then combinations of game state features provide information about how the important features interact with each other to influence sports outcomes. 

To demonstrate our work, one set of mimic trees is trained to predict this action-value for passes and shots in ice hockey and soccer. Our evaluation shows that our algorithms are computationally feasible (returning an answer in less than a day even on very large datasets) with great fidelity. Although we conduct experiments in sports, mimic learning with model trees is a general technique and can be applied to other domains. We also build mimic trees to predict the impact of these actions, which measures how much an action changes a team’s expected success.

Contributions: While mimic learning has been explored in machine learning \cite{DBLP:conf/nips/BaC14, DBLP:conf/amia/ChePKL16, DBLP:journals/tsmc/DanceyBM07}, to our knowledge it is new to sports analytics. Dense sports datasets can easily contain millions of data points. For example, our study uses a hockey dataset and a soccer dataset with more than seven million data points. As mentioned in section \ref{Computational Feasibility}, standard tree learning packages fail to process such large datasets. To address this severe computational challenge, we develop scalable model tree learning methods. The key is fast heuristic methods for finding promising thresholds for continuous predictor features (or co-variates). We also introduce a new data augmentation technique appropriate for counterfactual strategic settings.

% Paper Outline: We discuss previous related work, explain the dataset we used to conduct this work, and display two learned model trees as examples. We outline our main new contributions to model tree construction. The quantitative evaluation section shows the fidelity of our mimic learning method, interprets the mimic model through ranking observed features by their predictive importance and extracts rules that explain how the important features influence the outcome predictions.

\section{Previous Work}
Previous works on {\em mimic learning} have demonstrated that it is possible to learn a simple model, such as a shallow neural network \cite{DBLP:conf/nips/BaC14} or a tree-based model \cite{DBLP:journals/tsmc/DanceyBM07}, from an opaque complex model, such as a deep neural network, and maintain a similar predictive accuracy as the complex model. It has been shown that by doing so the prediction accuracy of the simple model outperforms the same simple model trained directly on the training set \cite{DBLP:conf/nips/BaC14}. Our work introduces three novel ideas that are important for action-value functions. (1) Whereas previous work uses simple regression trees for mimicking neural networks with continuous outputs, we use a linear model tree. Our experiments show that the additional expressive power of model trees compared to regression trees is essential for complex functions like expected success values in team sports. This agrees with the very recent work by \cite{DBLP:conf/ijcai/LiuS18}, who found that model trees are key for representing value functions in general reinforcement learning problems. (2) We investigate several fast heuristic methods
for building model trees. These heuristics are crucial for both computational feasibility
and fidelity. (3) We introduce action replacement, a new data augmentation technique for sports data. 

We apply mimic learning to construct interpretable models for action-value and impact functions. An action-value function, which is also called a Q-function, $Q(S_t,A_t)$ estimates  the expected future success of a team given the current match state $S_t$ and the current action $A_t$. For example, in the hockey model of \cite{schulte2017apples}, $Q(S_t,A_t)$ represents the conditional probability of a given team scoring the next goal given the event history (the state $S_t$) and the current action $A_t$. Other examples of action-value functions in sports analytics include expected points value (EPV) for basketball \cite{cervone2014pointwise}, expected possession value in soccer \cite{fernandez2019decomposing}, and expected points in NFL football \cite{yurko2019nflwar}. These studies have shown that action-values are a powerful way of valuing decisions and ranking players. However, the action-value function is not easy to interpret for sports stakeholders as it involves an expectation over future match trajectories. When the action-value function is estimated using neural nets, it is opaque to the user how it is computed \cite{fernandez2019decomposing, DBLP:conf/ijcai/LiuS18}. The combination of intransparency with usefulness makes the action-value function a suitable challenge for evaluating our mimic learning framework. 

\begin{figure*}[t]
  \centering
  \includegraphics[scale=0.3]{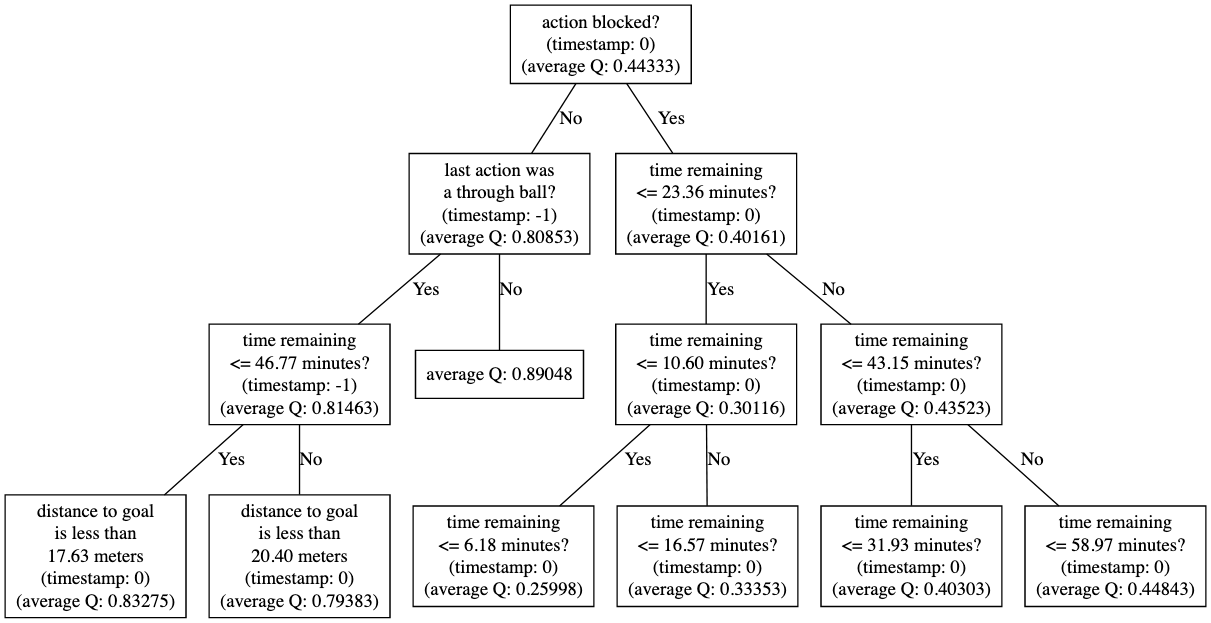}
  \caption{Model Tree Example With 4 Layers for Action-Value of Shots in Soccer}
  \label{fig:Model Tree Example with 4 Layers Soccer}
\end{figure*}

%Black-box interpretation systems. 
Mimic learning translates a black-box model into a white-box model. An alternative approach is to {\em analyze the neural net directly as a black box} \cite{guidottiblack}. A representative example of a black-box approach is Dalex \cite{DBLP:journals/jmlr/Biecek18}. Dalex utilizes different types of plots to visualize the behavior of a black-box model. For example, it performs residual diagnostics to analyze a regression model by drawing a plot that contains both the predictions of the model and their actual labels. Then, it is clear to spot the places where the model makes mistakes. Partial dependence plots show how the dependent variable changes if we change only one independent variable at a time. Also, Dalex and other explanation methods have been developed so far only for supervised regression and classification models \cite{DBLP:journals/jmlr/Biecek18}, not reinforcement learning. 
%While both white-box and black-box approaches have advantages for interpreting deep models, the main advantage of mimic learning is that is provides a comprehensive automated analysis of the complete knowledge acquired by the neural net, rather than relying on the user to select questions to pose to the black-box model.

We believe that converting a black-box model to a white-box model tree has two key advantages for sports analytics. (1) The model tree provides a comprehensive analysis of relevant interactions among domain variables. Interactions are represented in an intuitive visual tree format, so that even complex combinations of features remain comprehensible. (2) The mimic model can guide the user towards especially interesting and useful phenomena gleaned from the data. We illustrate this technique of ``mining the model" in our examples below.

\begin{figure*}[t]
  \centering
  \includegraphics[scale=0.3]{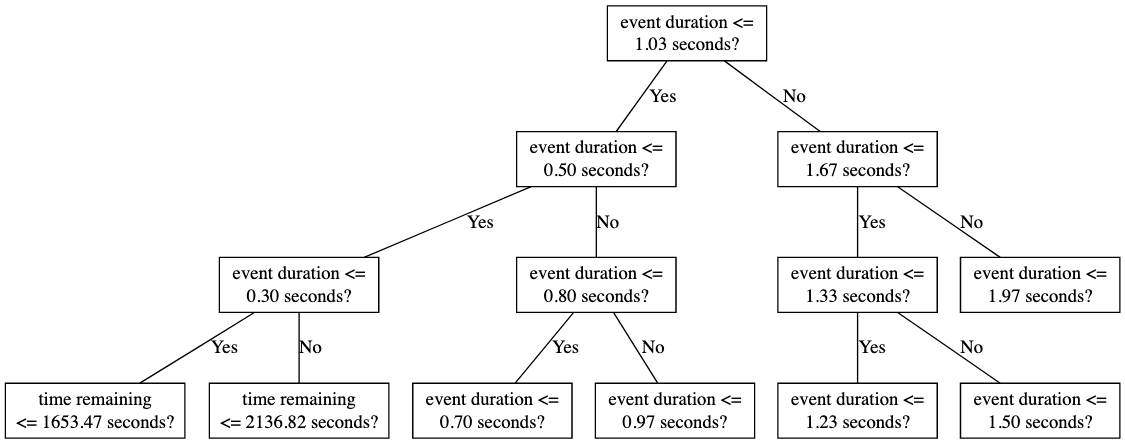}
  \caption{Model Tree Learned From a Biased DRL Model for Ice Hockey}
  \label{fig:Debug Neural Network Ice Hockey}
\end{figure*}

\section{Dataset and Neural Network Architecture}

\subsection{Input Features}
The data we used to conduct these experiments are collected by Sportlogiq. The data provides information about ice hockey game matches in the 2018-2019 NHL season and soccer game matches in the 2017-2018 season covering 10 leagues. Each data point represents a discrete event in a game, which combines information about the current situation and an action performed by a player on a team. Table \ref{tab:Independent Variables} lists all the input variables for ice hockey and soccer, and Figure \ref{fig:Ice Hockey Rink and Soccer Field} provides the visual demonstration of how coordinate systems are defined. In ice hockey, the x and y coordinates of puck are measured in feet from center ice, where -100 and 100 in the x-coordinate represent the planes at the backboards behind each net respectively, and -42.5 and 42.5 represent the planes at the boards at the sides with the players’ benches and penalty boxes respectively. The x coordinates on the defensive zone of a team are negative and that on the offensive zone are positive. In soccer, field length and width are evenly divided into 100 units, where coordinates (50, 50) represents the center spot of the field, (0, 50) and (100,50) represent the nets on the defensive zone and the offensive zone, respectively. In both ice hockey and soccer, the angle between the puck/ball and the goal is measured in radians clockwise from directly in front, such that $+\pi$, $-\pi$, $+\pi/2$, $-\pi/2$, are directly to the front, back, right and left of the net, respectively. The data also contain variables that specify actions and are normalized before being used for training.

\begin{table*}[t]
  \caption{Fidelity to Deep Model: RMSE on Test Set}
  \label{tab:RMSE}
  \begin{tabular}{l|p{1cm}|p{1cm}|p{1cm}|p{1cm}|p{1cm}|p{1cm}|p{1cm}|p{1cm}|}
  
  \cline{2-9}
  \multicolumn{1}{c}{} &
  \multicolumn{4}{|c|}{Ice Hockey} & \multicolumn{4}{|c|}{Soccer}\\
  \cline{2-9}
  \multicolumn{1}{c}{} &
  \multicolumn{2}{|c|}{Shots} & \multicolumn{2}{|c|}{Passes} & \multicolumn{2}{|c|}{Shots} & \multicolumn{2}{|c|}{Passes}\\

    \toprule
    Split methods & action-values & impacts & action-values & impacts & action-values & impacts & action-values & impacts \\
    \midrule
    Gaussian Mixture & 0.05483 & 0.01990 & 0.04276 & 0.00687 & 0.00698 & 0.01312 & 0.01000 & 0.00577\\
    \midrule
    Iterative Segmented Regression & 0.01441	& 0.01999 & \textbf{0.00964} & 0.00691 & \textbf{0.00508} & 0.01275 & \textbf{0.00997} & \textbf{0.00575}\\
    \midrule
    Sorting + Variance Reduction & \textbf{0.01219} & \textbf{0.01627} & 0.01012 & \textbf{0.00686} & 0.00646 & \textbf{0.01235} & 0.01092 & 0.00603\\
    \midrule
    Sorting + T-test & 0.05709 & 0.02487 & 0.06695 & 0.00935 & 0.01223 & 0.01377 & 0.01796 & 0.00597\\
    \midrule
    Null Model & 0.13924 & 0.05688 & 0.10808 & 0.01756 & 0.13648 & 0.11890 & 0.06151 & 0.00961\\
  \bottomrule
\end{tabular}
\end{table*}

\subsection{Target variable: Action Values and Impact Values}
% To generate ``soft" labels for the mimic model, we apply the temporal difference method SARSA \cite{rummery1994line, DBLP:conf/ijcai/LiuS18}, to learn a DRL model using a neural network representing an action-value function, which is also called a Q function in reinforcement learning.

To generate ``soft" labels for the mimic model, the neural net model outputs three action-values for each state and action pair $(S_t,A_t)$. The first action-value represents the probability of the home team having the next goal, the second action-value represents the probability of the away team having the next goal, and the third action-value represents the probability that the game ends before either team scores again. 

Another important quantity is action impact~\cite{DBLP:conf/ijcai/LiuS18}. Impact is defined as the difference between the action-value of a team given the current state-action pair and the action-value of the team given the previous state-action pair \[Impact(S_t,A_t) = Q(S_t,A_t) - Q(S_{t-1},A_{t-1}).\]
Impact represents the amount that an action performed by a player changes the probability of a given team scoring the next goal given the previous state. It is a useful refinement of action-values for measuring the importance of a specific action, by controlling for the general scoring chances of a team, which may not be under the control of the acting player. For example, in an empty net situation, the team driving towards the empty net has a high chance of scoring, which translates into a high action-value. But a player scoring on an empty net should not be given higher credit than for other goals. Therefore the previous works cited use the impact concept or a version of it to value actions and players, often called $<$metric$>$-added, e.g. EPV-added \cite{cervone2014pointwise, schulte2017apples, yurko2019nflwar}. In our evaluation, we carry out mimic learning for both the action-value and impact target variables. 

\subsection{Deep Reinforcement Learning Model}
Refer to \cite{DBLP:conf/ijcai/LiuS18}, the neural network architecture we use to construct the DRL model consists of five layers: an input layer, an LSTM hidden layer, two fully connected hidden layers and an output layer. Each hidden layer has 1000 ReLU neurons. Each game match is divided into episodes, such that each episode starts with either the beginning of a period or immediately after a team scoring a goal, and ends with either the end of a period or immediately when a team scoring a goal. We apply SARSA \cite{rummery1994line}, an on-policy temporal difference learning method ($\lambda = 1$), to the episodic dataset to estimate a Q-function. The parameters of the DRL model are optimized using minibatch gradient descent via Backpropagation Through Time with a fixed window-size of 10. The loss and update functions can be formulated as
\[ \mathcal{L}_{t} (\theta_{t}) = \mathbb{E} [(R_{t} + \hat{Q}(S_{t+1}, A_{t+1} ; \theta_{t}) - \hat{Q}(S_{t}, A_{t} ; \theta_{t}))^2 ] \]
\[ \theta_{t+1} = \theta_{t} - \alpha \cdot \nabla_{\theta} \mathcal{L}_{t} (\theta_{t}) \]
where $R_{t}$ is the reward at time step $t$, $\theta_{t}$ are parameter values at time step $t$, and $\alpha$ is the learning rate.

\subsection{Linear Model Tree Examples}
Figure \ref{fig:Model Tree Example with 4 Layers Ice Hockey} shows the first 4 layers of a shot impact model tree for ice hockey. To be consistent with the DRL model, the tree is also learned with a 10-step window of events preceding the current action, so predictor variables are shown with timestamps, where 0 indicates that the variable belongs to the same time $t$ as the current action $A_t$. As observations from the same time as the action are the most relevant to predicting its impact, the top layers of the tree split only on features with timestamp 0. The tree can be read in a top-down manner. The root node shows the first split condition and the average of the impact values in the training set. For each split, if the split condition is true, we follow the left edge to the next node; otherwise, we follow the right edge instead. For example, if the shot is blocked, then the tree checks the y-coordinate. If the shot occurred from more than 21.88 y-feet away, it checks the y-speed. Similarly, Figure \ref{fig:Model Tree Example with 4 Layers Soccer} shows the top 4 layers of a shot action-value (Q-value) model tree for soccer. The soccer tree also first splits on whether the shot is blocked or not. If the shot is blocked, then the tree checks if the last action before the shot was a through ball. For every child node, there is a new set of records assigned to the child node, and accordingly, a new average on every child node. When a leaf node is reached, a linear model is used to predict a target value. 
\[ \hat{y} = (\sum_{i} w_{i} \cdot x_{i}) + b \]
We can think of the conjunction of conditions along a branch as defining a discrete subset of the continuous input space \cite{uther1998tree}. 

\section{Model Tree Learning Outline}
In this section we outline our mimic learning method, emphasizing the novel contributions that support tree learning for sports analytics. We first describe our data augmentation, then the novel aspects of our method and how it supports learning interpretable trees. 
% The appendix provides a full description of our tree construction methods, including the pruning method. 
We provide our code available on-line~\footnote{\url{https://github.com/xiangyu-sun-789/Cracking-the-Black-Box-Distilling-Deep-Sports-Analytics}}.

\subsection{Data Augmentation}
An important strength of mimic learning is the ability to generate ``soft" labels for unobserved data points (sometimes called oracle coaching \cite{johansson2014accurate}) from the black-box model. This can be seen as a form of data augmentation. It is well-known that neural networks can be viewed as interpolating output labels \cite{DBLP:books/daglib/0087929}. Briefly, it can be shown that a trained neural network is equivalent to a kernel predictor (with a learned kernel) \cite{andras2002equivalence}, so labels assigned by the neural network are weighted averages of nearby data points. We introduce a new data augmentation technique in counterfactual strategic settings: asking the neural net to evaluate actions in settings where they do not usually occur in matches. We refer to this new data augmentation method tailored for action-functions as \textbf{action replacement}.

Given a target action $A'$, we randomly select an observed state-action pair $(S_t, A_t)$ where $A_t \neq A'$, and ask the neural network for a soft label $Q(S_t,A')$. For example, we may replace a sequence of events ending with a pass, by the same sequence ending in a shot. There are two benefits for action replacement. (1) It provides data for an action type across a wider set of situations than occurs in the data during professional play. Continuing the pass-to-shot example, predicting an action-value is equivalent to asking the neural network to evaluate the value of a player choosing to shoot rather than pass. (2) Because skilled players perform valuable actions in most situations, we expect that randomly altering actions receives a lower action-value. By exposing the mimic learner to data where the target action was not valuable, the tree model can learn which features distinguish match states that are favorable for an action. For example, shots are generally carried out close to the goal. By augmenting the data with low-value random shots from the neutral zone, the tree can learn the importance of shot distance as a feature. 

\subsection{Growing the Tree}
Trees are grown recursively. For any leaf node $l$, there is a set of data records that reach $l$. Following \cite{breiman2017classification}, our splitting criterion is to search for a predictive feature $x_i$, such that after splitting $l$ on $x_i$, the y-variance of the children is minimized. The main computational difficulty is that if $x_i$ is continuous, we need to find a breakpoint $c_i$ for splitting. The standard method for finding breakpoints for a potential split feature $x_i$ is to evaluate each $x_i$-value observed in the data. Evaluating each observed $x_i$-value raises severe computational difficulties because on a large dataset with a million or more records, there will typically be more than a million observed values for a continuous variable. Instead we introduce several fast heuristics for identifying promising breakpoints $c_i$, described in section \ref{Computing Split Points}. Splits are restricted such that every child node is assigned at least $m = 100$ data records. By increasing the sample size $m$, the user can obtain a smaller tree but with less fidelity. The appendix provides further implementation details.

\begin{table*}[t]
  \caption{Independent Variables}
  \label{tab:Independent Variables}
  \begin{tabular}{l|l|p{2cm}}
    \toprule
    Variables for Ice Hockey & Type & Range \\
    \midrule
    time remaining in seconds& continuous & [0, 3600]\\
    \midrule
    x coordinate of puck & continuous & [-100, 100]\\
    \midrule
    y coordinate of puck & continuous & [-42.5, 42.5]\\
    \midrule
    score differential & discrete & ($-\infty$, $+\infty$)\\
    \midrule
    manpower situation & discrete & \{even strength, short handed, power play\}\\
    \midrule
    action blocked & discrete & \{true, false, undetermined\}\\
    \midrule
    x velocity of puck & continuous & ($-\infty$, $+\infty$)\\
    \midrule
    y velocity of puck & continuous & ($-\infty$, $+\infty$)\\
    \midrule
    event duration & continuous & [0, $+\infty$)\\
    \midrule
    angle between puck and goal & continuous & [$-\pi$, $+\pi$]\\
    \midrule
    home team taking possession  & discrete & \{true, false\}\\
    \midrule
    away team taking possession & discrete & \{true, false\}\\
    \midrule
    action & discrete & one-hot for 27 actions\\
  \bottomrule
\end{tabular}\hfill
\begin{tabular}{l|l|p{2cm}}
    \toprule
    Variables for Soccer & Type & Range \\
    \midrule
    time remaining in minutes & continuous & [0, 100]\\
    \midrule
    x coordinate of ball & continuous  & [0, 100]\\
    \midrule
    y coordinate of ball & continuous  & [0, 100]\\
    \midrule
    distance to goal in meters & continuous & [0,110] \\
    \midrule
    score differential & discrete & (-$\infty$, +$\infty$)\\
    \midrule
    manpower situation & discrete  & [-5, 5] \\
    \midrule
    action blocked & discrete & \{true, false\}\\
    \midrule
    x velocity of ball & continuous & (-$\infty$, +$\infty$)\\
    \midrule
    y velocity of ball & continuous & (-$\infty$, +$\infty$)\\
    \midrule
    event duration & continuous  & [0, +$\infty$) \\
    \midrule
    angle between ball and goal & continuous & [$-\pi$, $+\pi$]\\
    \midrule
    home team taking possession  & discrete & \{true, false\}\\
    \midrule
    away team taking possession & discrete & \{true, false\}\\
    \midrule
    action & discrete & one-hot for 43 actions \\
  \bottomrule
\end{tabular}
\end{table*}

\subsection{Heuristics for Computing Split Points} \label{Computing Split Points}

We refer to the group of data points with $x_i \leq c_i$ and $x_i > c_i$ as the split groups. We investigated several fast heuristic methods for selecting promising breakpoints $c_i$ for a given input feature $x_i$. These heuristics are crucial for both computational feasibility and fidelity. The key idea behind our methods is to sort all the data points by their $x_i$-value, then choose a breakpoint $c_i$ that maximizes the difference in the $y$-distributions of the datapoint groups created by $c_i$. Our proposed heuristics combine sorting with variance reduction and t-test, or use segmented regression with efficient iterative estimation as a subroutine to achieve fast performance on large datasets. We apply heuristic with Gaussian Mixture as our baseline.

\subsubsection{Sorting with Variance Reduction}
Maximizing the difference in the $y$-distributions of the datapoint groups after a split can be estimated by variance reduction on $y$. Simply sorting first on $x_i$ allows us to incrementally estimate the variance reduction for every $x_i$-value quickly with a single pass through the dataset, as shown by the following equations:
\begin{align}
    \sigma^2 & = \frac{1}{N_{1}} \cdot \sum^{N_{1}}_{n=1}(y_n-\mu)^2 \nonumber \\ 
    % & = \frac{1}{N_{1}} \cdot (\sum^{N_{1}}_{n=1}{y_n}^2 - 2 \cdot \mu \cdot \sum^{N_{1}}_{n=1} y_n + {N_{1}} \cdot \mu^2) \nonumber \\
    % & = \sum^{N_{1}}_{n=1} \frac{{y_n}^2}{N_{1}} - 2 \cdot \mu \cdot \sum^{N_{1}}_{n=1} \frac{y_n}{N_{1}} + {N_{1}} \cdot \frac{1}{N_{1}} \cdot \mu^2 \nonumber \\
    & = (\sum^{N_{1}}_{n=1} \frac{{y_n}^2}{N_{1}}) - (\sum^{N_{1}}_{n=1}\frac{y_n}{N_{1}})^2 \label{eq1}
\end{align}
where $N_{1}$ represents the data points in one split group after splitting on an $x_i$-value, $\mu$ is the $y$ mean of the split group. Both terms in equation \ref{eq1} are calculated incrementally in a single pass for all $x_i$-values.

\subsubsection{Sorting with T-test}
This method also sorts all the data points by their $x_i$-value. Then, it uses the test-statistic of two-sample Welch’s t-test \cite{lee1992optimal} to evaluate breakpoints. 
\[ \text{t-score} = \frac{\mu_1 - \mu_2}{\sqrt{\frac{{\sigma_1}^2}{N_1} + \frac{{\sigma_2}^2}{N2}}} \]
The t-test measures the y-difference between the two split groups sperated by an $x_i$-value. We select the $x_i$-value that produces the largest t-score as breakpoint $c_i$. As with variance reduction, the t-score can also be computed incrementally in linear time.

\subsubsection{Iterative Segmented Regression}
Segmented regression performs a piecewise linear regression of y on $x_i$ with a breakpoint $c_i$ between two line segments \cite{vens2006simple}. 
\begin{equation*}
  \hat{y}=\begin{cases}
    \alpha \cdot x_i & \text{for $x_i \leq c_i$}\\
    (\alpha + \beta) \cdot x_i - \beta \cdot c_i & \text{for $x_i > c_i$}
  \end{cases}
\end{equation*}
We first use segmented regression as a subroutine with an efficient iterative approach \cite{muggeo2003estimating} to find a breakpoint candidate on each feature $x_i$. The following algorithm elaborates on the iterative approach for a feature $x_i$ at iterative step $s$:
\begin{enumerate}
    \item 
        \begin{equation*}
            U^{s} = \begin{cases}
                x_i - c_{i}^{s} & \text{for $x_i > c_{i}^{s}$}\\
                0 & \text{otherwise}
            \end{cases}
        \end{equation*}
        \begin{equation*}
            V^{s} = \begin{cases}
                -1 & \text{for $x_i > c_{i}^{s}$}\\
                0 & \text{otherwise}
            \end{cases}
        \end{equation*}
    \item fit the model \[ \hat{y} = \alpha \cdot x_i + \beta \cdot U^{s} + \gamma \cdot V^{s} \]
    \item update the breakpoint $c_{i}$ \[ c_{i}^{s+1} = \frac{\gamma}{\beta} + c_{i}^{s}\]
    \item repeat the process until the breakpoint $c_{i}$ is converged or the maximum iterative step is reached.
\end{enumerate}
Then, for each breakpoint candidate, we calculate the y-variances of two groups separated by the breakpoint candidate. We select the breakpoint candidate that maximizes the difference in the y-variances of the two split groups as the breakpoint $c_i$.

\begin{figure*}[t]
  \subfigure{
  \includegraphics[width=\linewidth]{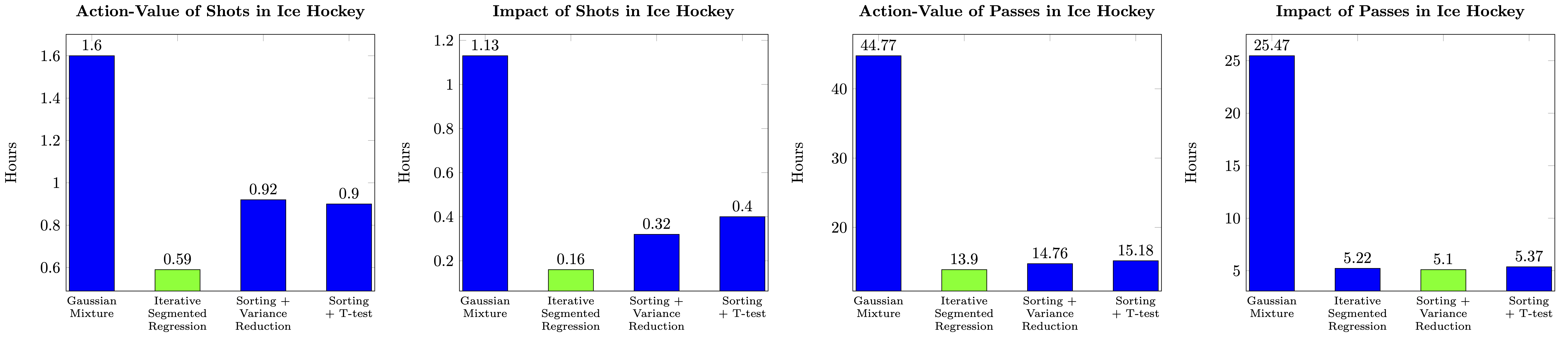}
  }
  \subfigure{
  \includegraphics[width=\linewidth]{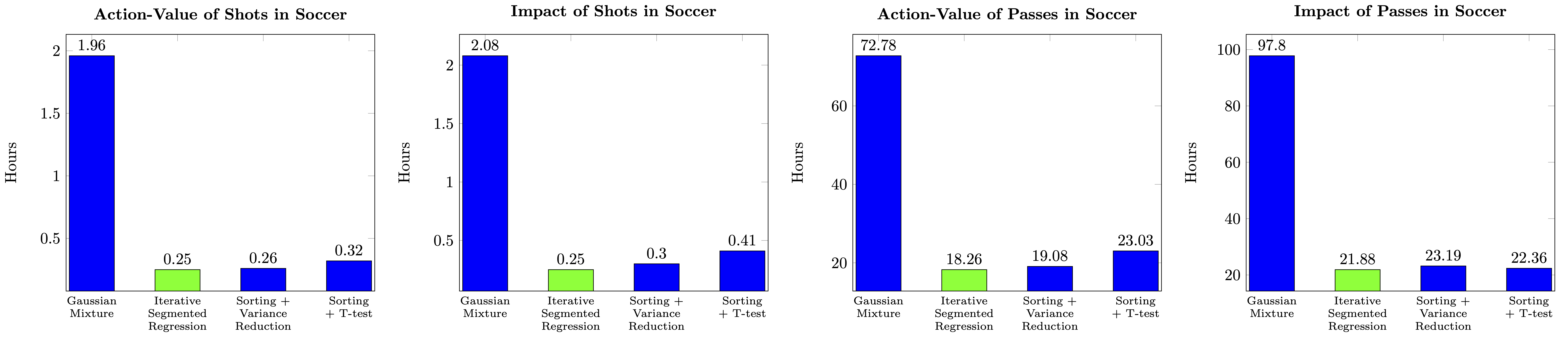}
  }
  \caption{Running Time}
  \label{fig:Running Time}
\end{figure*}

\begin{figure}[h]
  \centering
  \includegraphics[scale=0.4]{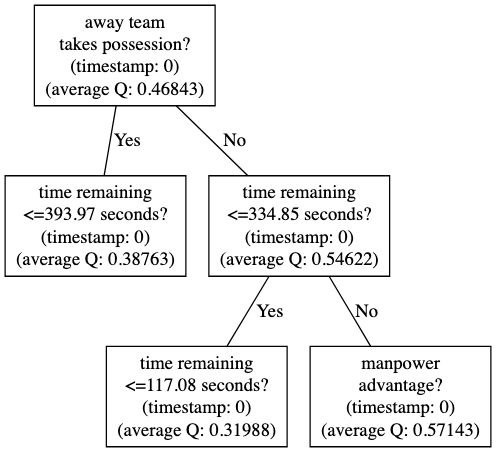}
  \caption{Rule Example 1 for Action-Value of Shots in Ice Hockey. The model tree for ice hockey produces a prediction for the Q-probability that the home team scores the next goal after a shot.}
  \label{fig:Rule Example 1 for Action-value of Shots in Ice Hockey}
\end{figure}

\begin{figure}[h]
  \centering
  \includegraphics[scale=0.4]{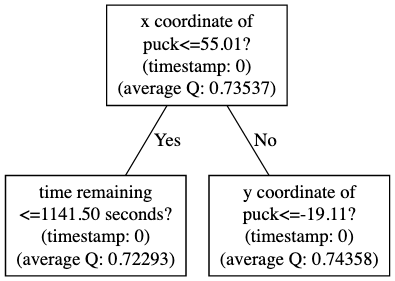}
  \caption{Rule Example 2 for Action-Value of Shots in Ice Hockey}
  \label{fig:Rule Example 2 for Action-value of Shots in Ice Hockey}
\end{figure}

\subsubsection{Gaussian Mixture}
This method uses the expectation maximization algorithm to calculate a two-component bivariate Gaussian mixture model \cite{dobra2002secret} for $(x_i,y)$ data pairs.
\[p(x_i, y) = \sum^{2}_{k=1} \pi_{k} \cdot \mathcal{N}(x_i, y | \mu_{k}, \Sigma_{k}) \]
Then, the breakpoint $c_i$ that best separates the two Gaussian clusters on each predictor variable $x_i$ can be computed in closed form by quadratic discriminant analysis.

\section{Evaluation}
Here, we evaluate the mimic-learned models' fidelity, that is, their ability to match the output of the black-box DRL model. We also rank features for predicting shot action-values and impacts by importance, then show rules that describe how the important features influence the predictions. All three of Sorting with Variance Reduction, Sorting with T-test and Iterative Segmented Regression are fast enough for scalable model tree learning, with Iterative Segmented Regression as the fastest method. Details on computational costs can be found in Figure \ref{fig:Running Time}.

\subsection{Fidelity}
A mimic model must show strong fidelity \cite{DBLP:journals/tsmc/DanceyBM07}, that is, the root mean squared difference (RMSE) between the prediction of the tree and the prediction of the DRL model must be small.

As Table ~\ref{tab:RMSE} shows, Iterative Segmented Regression and Sorting with Variance Reduction achieve greater fidelity on test set than other methods. Given its speed (Figure \ref{fig:Running Time}), we recommend Iterative Segmented Regression as a good default method, and Sorting with Variance Reduction as a close second. The null model calculates the mean value of the response variable and uses that as its prediction. Table \ref{tab:Correlation} reports high correlations between the outputs of the neural and mimic models: for iterative segmented regression, they are almost always above 0.9 and in many cases above 0.99.

\subsection{Feature Importance}
A basic question for understanding a neural net is which input features most influence its predictions. Given a model tree, we can compute the feature importance as the sum of variance reductions over all splits that use the feature \cite{DBLP:conf/ijcai/LiuS18}. Table \ref{tab:Top 10 Features for Shots} shows the feature importance of the top 10 most relevant features for the action-value of shots in ice hockey and soccer, with feature frequency defined as how many times the tree splits on the feature. Time remaining is important for both ice hockey and soccer because the probability of either team scoring another goal decreases quickly when not much time is left. Moreover, time remaining has a stronger influence on ice hockey than soccer because there are generally more goals in ice hockey. At the beginning of a game, the probability of a team scoring in ice hockey is higher than that in soccer. As time goes towards the end, the probability of scoring in ice hockey decreases more quickly than that in soccer. Unsurprisingly, puck/ball to goal distance and action outcomes (i.e. shots being blocked or not) are also among the most relevant features for shots in ice hockey and soccer.

\begin{table*}[t]
  \caption{Top 10 Features for Shots}
  \label{tab:Top 10 Features for Shots}
  \begin{tabular}{p{4.5cm}|p{1.5cm}|p{1.5cm}}
    \toprule
    Ice Hockey & Feature \newline Importance & Feature \newline Frequency\\
    \midrule
    time remaining ($t_{0}$) & 0.0594 & 248\\
    y coordinate of puck ($t_{0}$) & 0.03418 & 228\\
    x coordinate of puck ($t_{0}$) & 0.02646 & 153\\
    action blocked ($t_{0}$) & 0.02016 & 12\\
    manpower situation ($t_{0}$) & 0.01203 & 14\\
    home ($t_{0}$) & 0.00629 & 1\\
    angle between puck and goal ($t_{0}$) & 0.00164 & 32\\
    time remaining ($t_{-1}$) & 0.00072 & 9\\
    action: reception ($t_{-1}$) & 0.00061 & 5\\
    score differential ($t_{-1}$) & 0.00026 & 23\\
  \bottomrule
  \end{tabular}\hfill
  \begin{tabular}{p{4cm}|p{1.5cm}|p{1.5cm}}
    \toprule
    Soccer & Feature \newline Importance & Feature \newline Frequency\\
    \midrule
    action blocked ($t_{0}$) & 0.01524 & 1\\
    time remaining ($t_{0}$) & 0.00711 & 36\\
    distance to goal ($t_{0}$) & 0.00144 & 31\\
    action: through ball ($t_{-1}$) & 0.00079 & 1\\
    event duration ($t_{0}$) & 0.00068 & 8\\
    time remaining ($t_{-1}$) & 0.00059 & 12\\
    y velocity of ball ($t_{0}$) & 0.00036 & 5\\
    x coordinate of ball ($t_{0}$) & 0.00015 & 28\\
    manpower situation ($t_{0}$) & 0.00011 & 1\\
    action: cross ($t_{-1}$) & 0.00011 & 2\\
  \bottomrule
\end{tabular}
\end{table*}

\subsection{Rule Extraction}
We can extract rules that can be easily interpreted by humans from a model tree. The rules highlight relevant interactions among input features. They also expand on the feature importance by showing how the important features influence the predictions of the neural network.

\begin{figure}[h]
  \centering
  \includegraphics[scale=0.4]{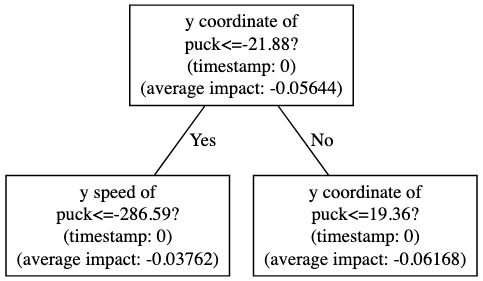}
  \caption{Rule Example for Impact of Shots in Ice Hockey}
  \label{fig:Rule Example for Impact of Shots in Ice Hockey}
\end{figure}

For shots in ice hockey, Figure \ref{fig:Rule Example 1 for Action-value of Shots in Ice Hockey} is a part of a tree to demonstrate how rules can be extracted. First, how good a shot is for the home team is related to which team is taking possession of the puck. In other words, whether the shot is performed by the home team or the away team. By looking at the average Q-values of the corresponding child nodes, we see that it is better for the home team if they take a shot than if the away team takes a shot. If the shot is by the home team, its Q-values are related to the time remaining in the game: with little time left (less than 335 seconds), there is less of a chance of any team scoring. However, given sufficient time, the next feature the tree considers is whether the home team has a manpower advantage. Figure \ref{fig:Rule Example 2 for Action-value of Shots in Ice Hockey} shows another part of the same tree. It supports the rule that the action-value of shots in ice hockey is better when the puck is closer to the net (recall the defensive zone has negative x coordinates and offensive zone has positive x coordinates). If the puck is sufficiently close, then the tree next considers the y-coordinate of the puck location. Figure \ref{fig:Rule Example for Impact of Shots in Ice Hockey} is an excerpt from Figure \ref{fig:Model Tree Example with 4 Layers Ice Hockey} after a shot is blocked. It extracts the rule of impacts such that when a shot is blocked by the opposite team, the impact of the action is less bad when the puck is far from the net. If the puck is close to the net when the shot is blocked, a good opportunity to a goal is lost, therefore, the impact is much worse.

\begin{figure}[h]
  \centering
  \includegraphics[scale=0.4]{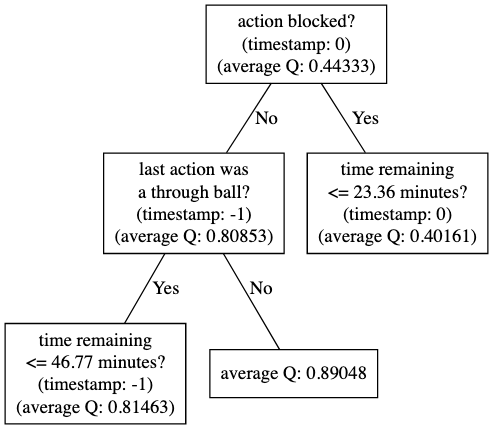}
  \caption{Rule Example for Action-Value of Shots in Soccer}
  \label{fig:Rule Example for Action-value of Shots in Soccer}
\end{figure}

\begin{figure}[h]
  \centering
  \includegraphics[scale=0.4]{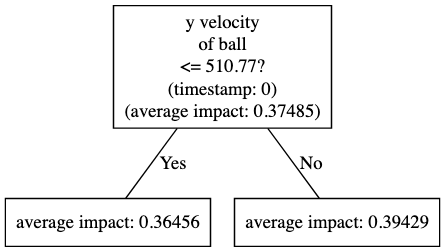}
  \caption{Rule Example for Impact of Shots in Soccer}
  \label{fig:Rule Example for Impact of Shots in Soccer}
\end{figure}

For shots in soccer, Figure \ref{fig:Rule Example for Action-value of Shots in Soccer} is the top part of Figure \ref{fig:Model Tree Example with 4 Layers Soccer}. As in ice hockey, it shows that action-values of shots are better when shots are not blocked. Furthermore, it gives an insight that through-ball passes are not the best thing to do to assist a goal. The tree suggests that if a shot is taken right after a through-ball pass, the shot is usually less promising to a goal. For the impact of shots in soccer, Figure \ref{fig:Rule Example for Impact of Shots in Soccer} presents a rule such that shots that are fast in y-direction usually have high impact values.

\subsection{Debugging Deep Neural Networks}
Because of the transparency of tree-based models, the tree learned from a DRL model can highlight potential problems in the DRL model. Figure \ref{fig:Debug Neural Network Ice Hockey} is part of a tree learned from an early version of a DRL Q-function model for ice hockey. The tree splits frequently on the feature Event Duration. When we presented the tree to ice hockey experts, the splits drew their attention - splitting frequently on duration conflicts with their expertise. As a consequence, we discovered an information leakage introduced in the data processing that extracted the duration feature, which caused it to highly correlate with Q-values. Without an interpretable model such as the tree, it is almost impossible to spot the spurious behaviour from the black box of the deep neural network.
%The tree assisted the experts to understand the DRL model and discover the bias.}

\section{Computational Feasibility} \label{Computational Feasibility}
Several standard model tree learning packages failed to build on our large dataset due to their memory limitations. These include pyFIMTDD \cite{ikonomovska2011learning} and production systems such as Weka \cite{weka} and GUIDE \cite{guide} that have been deployed in commercial applications. This highlights the need for new computational methods that can extend to large sports datasets. 

All the experiments were performed on a computing node, provided by Compute Canada, with 4 core CPU and 64GB of RAM. Figure \ref{fig:Running Time} shows the running time in hours for two actions, shots (150K events) and passes (1M events), for each breakpoint heuristic.

Iterative Segmented Regression is the fastest method. The sorting methods are slower but still bring the computational cost of the analysis to less than a day. 

\section{Conclusion and Future Work}

The predictions of trained models must be explained if sports experts are to benefit fully from modern machine learning. Learning to mimic a neural net with a linear model tree offers a sweet spot in the accuracy-transparency trade-off: accurate predictions with rules and features that explicate the insights gained from data analysis. We introduced a new action replacement technique for augmenting sports data with soft labels from the neural network. Another new contribution are fast new heuristic methods for model tree construction that scale to large datasets. Mimic learning allows sports analytics to combine the predictive power of modern machine learning techniques with explanations and actionable insights for sports experts. A direction for future work is to investigate whether model trees support transfer learning between sports, as in our example of blocked shots. While specific threshold may be domain specific, trees can identify which combinations of features are important across sports.

\bibliographystyle{ACM-Reference-Format}
\bibliography{main}

\clearpage
\appendix

\section{Dataset and Model Tree Construction}
\subsection{Predictive Variables in the Dataset}
For ice hockey, we only considered regulation time for each game match and did not consider overtime periods (which are governed by substantially different rules). 

\subsection{Tree Construction}
When describing tree learning algorithms, we use $x=(x_1,x_2,\dots,x_m)$ for input features (covariates), and y for the output (dependent) variable. In our application, $x$ is a feature set for a state $S$, and $y=Q(S,A)$ is the output Q-value of the neural net for a fixed action $A$.  The standard schema for growing a model tree is as follows \cite{breiman2017classification}.

\begin{enumerate}
  \item \textit{Initialization}. Start with the root node. Assign all data records to it.
  \item \textit{Growth Phase}. At every leaf node $l$, for every input feature $x_i$, compute a promising breakpoint $c_i$.
  \begin{enumerate}
      \item If no split $(x_i,c_i)$ improves the splitting criterion, keep $l$ as a leaf node.
      \item Otherwise find the split $(x_i,c_i)$ that maximizes the splitting criterion. Assign the data records for $l$ with $x_i \leq c_i$ to the left child of $l$, and those with $x_i > c_i$ to the right child of $l$. 
  \end{enumerate}
  \item \textit{Pruning Phase}. Consider the parent $v$ of two leaf nodes $v_1$ and $v_2$. If the pruning criterion improves by replacing the two leaf nodes by $v$ as the leaf, prune the two leaf nodes.
\end{enumerate}

\subsection{Splitting Criterion}
Following \cite{breiman2017classification}, we split the tree at the point $(x_i,c_i)$ that gives the greatest reduction in y-variance, so the splitting criterion is:
\[\text{Variance}(s) - [\frac{N_{s_t}}{N_s} \cdot \text{Variance}(s_t) + \frac{N_{s_f}}{N_s} \cdot \text{Variance}(s_f)]\]
where $s$ is the whole set of data records $(x_i)$ on a node, $s_t$ is the set of data records on a child node for which the split condition is true $(x_i \leq c_i)$, $s_f$ is the set of data records on another child node for which the split condition is false $(x_i > c_i)$, and $N_s$ represents the number of data records in set $s$.

\subsection{Pruning the Tree}
Growing a tree by variance reduction captures many informative interactions but tends to overfit. It is therefore necessary to add a pruning phase. The dual objectives of the pruning phase are to maximize the fidelity of the tree and reduce its complexity to increase interpretability. This trade-off can be expressed as a regularized linear regression with a complexity penalty on the weight parameters. For a tree node $v$, let $N_v$ be the number of data records assigned to $v$. The loss function at node $v$ is given by 
\[E_v = \text{arg} \min_w \sum^{N_v}_{j=1} y_j - (w \cdot x_j) + \lambda \cdot R(w)\] 
A split at node $v$ is removed if doing so decreases the $E$ value for node $v$. For the complexity penalty R we use the L0-norm (number of parameters) or the L1-norm (ridge regression). In our experiments, the L0-norm gives a smaller tree and the L1-norm gives better fidelity on the held-out testing set. By increasing the trade-off parameter $\lambda$, the user can obtain a smaller tree but with less fidelity. 

\subsection{Pruning Criterion}
For a tree node $v$, let $N_v$ be the number of data records assigned to $v$. Consider the parent $v$ of two leaf nodes $v_1$ and $v_2$. The pruning criterion is $E$, so if $E_v< E_{v1} + E_{v2}$, then we prune the two leaf nodes and make $v$ a new leaf node. Pruning is repeated until $E_v>= E_{v1} + E_{v2}$ for all leaf node parents $v$.

\begin{table*}[t]
  \caption{Correlation Between Predictions of Model Tree and DRL Model}
  \label{tab:Correlation}
  \begin{tabular}{l|p{1cm}|p{1cm}|p{1cm}|p{1cm}|p{1cm}|p{1cm}|p{1cm}|p{1cm}|}
  
  \cline{2-9}
  \multicolumn{1}{c}{} &
  \multicolumn{4}{|c|}{Ice Hockey} & \multicolumn{4}{|c|}{Soccer}\\
  \cline{2-9}
  \multicolumn{1}{c}{} &
  \multicolumn{2}{|c|}{Shots} & \multicolumn{2}{|c|}{Passes} & \multicolumn{2}{|c|}{Shots} & \multicolumn{2}{|c|}{Passes}\\

    \toprule
    Split methods & action-values & impacts & action-values & impacts & action-values & impacts & action-values & impacts \\
    \midrule
    Gaussian Mixture & 0.91498 & 0.93709 & 0.94737 & 0.91687 & 0.99458 & 0.99001 & 0.98386 & 0.81639 \\
    \midrule
    Iterative Segmented Regression & 0.99436 & 0.93620 & \textbf{0.99601} & 0.92018 & \textbf{0.99650} & 0.99422 & \textbf{0.98695} & \textbf{0.81966} \\
    \midrule
    Sorting + Variance Reduction & \textbf{0.99593} & \textbf{0.95834} & 0.99561 & \textbf{0.92137} & 0.99480 & \textbf{0.99459} & 0.98438 & 0.80024 \\
    \midrule
    Sorting + T-test & 0.91036 & 0.89935 & 0.79761 & 0.85017 & 0.98943 & 0.98705 & 0.95690 & 0.79132 \\
    \midrule
    Null Model & 0.00000 & 0.00000 & 0.00000 & 0.00000 & 0.00000 & 0.00000 & 0.00000 & 0.00000 \\
  \bottomrule
\end{tabular}
\end{table*}

\end{document}